\documentclass[letterpaper]{article} 
\usepackage{aaai2026}  
\usepackage{times}  
\usepackage{helvet}  
\usepackage{courier}  
\usepackage[hyphens]{url}  
\usepackage{graphicx} 
\usepackage{afterpage}
\usepackage{natbib}  
\usepackage{caption} 
\usepackage{algorithm}

\usepackage{newfloat}
\usepackage{listings}
\DeclareCaptionStyle{ruled}{labelfont=normalfont,labelsep=colon,strut=off} 
\floatstyle{ruled}
\newfloat{listing}{tb}{lst}{}
\floatname{listing}{Listing}
\usepackage{aaai2026}  
\usepackage{times}  
\usepackage{helvet}  
\usepackage{courier}  
\usepackage[hyphens]{url}  
\usepackage{graphicx} 
\usepackage{subcaption} 
\usepackage{multirow}
\usepackage{natbib}  
\usepackage{caption} 
\usepackage{xcolor}  
\usepackage{colortbl}  

\usepackage{amsmath,amsfonts}
\usepackage{amsthm}
\usepackage{algorithm}
\usepackage{algpseudocode}
\usepackage{booktabs} 	
\usepackage{graphicx}

\usepackage{newfloat}
\usepackage{listings}
\DeclareCaptionStyle{ruled}{labelfont=normalfont,labelsep=colon,strut=off} 
\floatstyle{ruled}
\newfloat{listing}{tb}{lst}{}
\floatname{listing}{Listing}
\usepackage{amsfonts}
\usepackage{amsmath}
\usepackage{xcolor}

\definecolor{mytop}{RGB}{143,188,143}
\definecolor{mysec}{RGB}{190,215,103}
\definecolor{mythird}{RGB}{250,255,100}

\title{Constrained Particle Seeking: Solving Diffusion Inverse Problems
with \\ Just Forward Passes}
\author{
    Hongkun Dou\textsuperscript{\rm 1}\equalcontrib, Zike Chen\textsuperscript{\rm 1}\equalcontrib, Zeyu Li\textsuperscript{\rm 1}, Hongjue Li\textsuperscript{\rm 1} \thanks{Corresponding authors.}, Lijun Yang\textsuperscript{\rm 1}, Yue Deng\textsuperscript{\rm 1}\textsuperscript{\rm 2}
    \\
}
\affiliations{
    \textsuperscript{\rm 1}Beihang University \\
    \textsuperscript{\rm 2}Beijing Zhongguancun Academy\\
        \{douhk, zy2415215, lizeyu123478, lihongjue, yanglijun, ydeng\}@buaa.edu.cn.

}

\usepackage{bibentry}

\begin{document}

\maketitle

\begin{figure*}[h!]
     \centering
\includegraphics[width=0.98\textwidth]{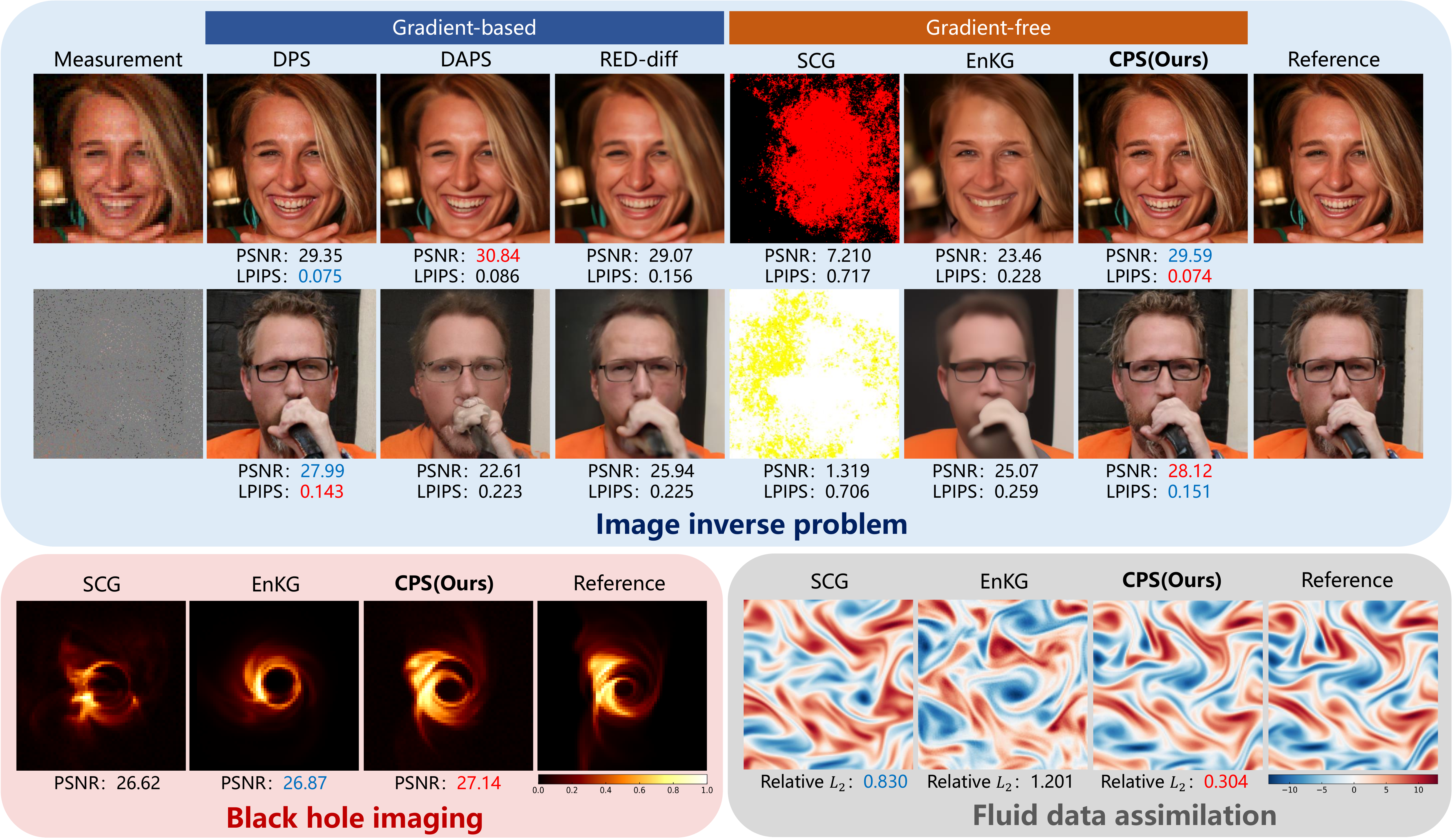}
    \caption{Qualitative comparison between CPS and baseline methods. The proposed CPS relies solely on the forward pass of the observation model to solve both image and scientific inverse problems. It performs competitively with advanced gradient-based methods and significantly outperforms gradient-free methods. The best results are highlighted in \textcolor{red}{red},  and the second-best results are shown in \textcolor[rgb]{0.25,0.5,0.80}{blue}.}
\end{figure*}

\begin{abstract}
Diffusion models have gained prominence as powerful generative tools for solving inverse problems due to their ability to model complex data distributions. However, existing methods typically rely on complete knowledge of the forward observation process to compute gradients for guided sampling, limiting their applicability in scenarios where such information is unavailable. In this work, we introduce \textbf{\emph{Constrained Particle Seeking (CPS)}}, a novel gradient-free approach that leverages all candidate particle information to actively search for the optimal particle while incorporating constraints aligned with high-density regions of the unconditional prior. Unlike previous methods that passively select promising candidates, CPS reformulates the inverse problem as a constrained optimization task, enabling more flexible and efficient particle seeking. We demonstrate that CPS can effectively solve both image and scientific inverse problems, achieving results comparable to gradient-based methods while significantly outperforming gradient-free alternatives.  Code is available at \url{https://github.com/deng-ai-lab/CPS}.
\end{abstract}


\section{Introduction}
The inverse problem is a fundamental challenge across various fields, including computational imaging \citep{tonolini2020variational,mccann2017convolutional}, medicine \citep{song2021solving}, geophysics \citep{schwarzbach2011three}, and fluid dynamics \citep{li2020fourier}. Its goal is to recover the original signal, $\boldsymbol{x} \in \mathbb{R}^d$, from indirect and noisy observations, $\boldsymbol{y} \in \mathbb{R}^m$, typically modeled as:
\begin{equation}
\boldsymbol{y} = \mathcal{H}(\boldsymbol{x}) + \boldsymbol{\eta}    
\end{equation}
where $\mathcal{H}(\cdot): \mathbb{R}^d \rightarrow \mathbb{R}^m$ represents the forward measurement operator and $\boldsymbol{\eta}$ denotes additive noise. This problem is inherently ill-posed ($m < d$), meaning multiple potential solutions may satisfy the given observations. To resolve this, appropriate priors and regularization techniques are essential to mitigate ill-posedness and ensure a meaningful solution.

Recent advances in generative modeling, particularly diffusion models \citep{ho2020denoising,song2020score, lipman2022flow, rombach2022high}, have proven effective as plug-and-play priors for solving inverse problems, offering significant advantages over traditional manual priors \citep{tibshirani1996regression,iordache2012total}. Existing approaches rely on gradients of the observation process or pseudo-inverse information to guide the unconditional sampling process, ensuring that the generated results align with observations \citep{zhang2025improving,wang2022zero,chung2022diffusion, kawar2022denoising,mardani2023variational}. However, this dependency on privileged information limits the applicability of these methods, especially in real-world scenarios. For example, in some scenarios, forward models often involve numerical simulations that are highly nonlinear or computationally expensive, making gradient computation challenging \citep{oliver2008inverse,gillhofer2019gan, duong2020vec2face,kawar2022jpeg}.

Several efforts have explored the use of pre-trained diffusion models to solve inverse problems with black-box access to the forward model. One such approach is SCG \citep{huang2024symbolic}, which employs a strategy akin to rejection sampling. At each step of the reverse denoising process, SCG samples multiple candidate particles and retains only the one that best aligns with the observation, discarding the others. While this method is simple and intuitive, it suffers from inefficiency due to the discarding of many candidates at each step. Another notable method, EnKG \citep{zhengensemble}, is inspired by ensemble Kalman filtering \citep{katzfuss2016understanding}. EnKG initializes and maintains a set of particles to represent the state, progressively refining it based on observations. However, EnKG requires maintaining thousands of particles in high-dimensional inverse problems, resulting in significant computational overhead. Additionally, DPG \citep{tang2024solving} avoids the calculation of derivatives by using policy gradients, but it necessitates adding noise perturbations to clean samples, which can push it outside the data manifold. More importantly, these methods often underperform relative to gradient-based approaches when gradients are available.

To overcome these limitations, we propose \textbf{\emph{Constrained Particle Seeking (CPS)}}, a principled approach that synthesizes information from all candidate particles to identify an optimal solution. We argue that each candidate particle, rather than being discarded, offers valuable local insights that can collectively guide the denoising process. In CPS, we reframe the diffusion inverse problem as a constrained optimization task. The objective is to find a particle that optimally aligns with the observation, where the alignment is measured by a local surrogate model of the forward measurement. This surrogate is robustly estimated using the full set of candidate particles. Concurrently, we introduce a constraint that ensures the resulting particle remains within the high-probability density region of the reverse diffusion process, thereby preserving consistency with the diffusion prior. By leveraging all available candidates, CPS enhances sampling efficiency and achieves a superior balance between observation fidelity and prior knowledge. Furthermore, we incorporate a \emph{Restart} strategy \citep{xu2023restart,lugmayr2023inpainting} to iteratively correct for the cumulative errors inherent in the sequential sampling process, which significantly improves the robustness and accuracy of the final solution.

We evaluated CPS on various tasks, including benchmark image inverse problems and two scientific inverse problems: black hole imaging and fluid data assimilation, both characterized by highly nonlinear measurement processes. Our results show that CPS not only matches the performance of popular gradient-based methods in image inverse problems but also significantly enhances efficiency in forward processes where gradients are unavailable.

\section{Preliminaries and Task Formulation}

\subsection{Diffusion Models}
The diffusion model \citep{ho2020denoising,song2020score} defines a stochastic process $\{\boldsymbol{x}_t\}_{t \in [0,T]}$ that progressively transforms data $\boldsymbol{x}_0 \sim p_{\text{data}}$ into noise. This process is governed by the following differential equation:
\begin{equation}\label{forward}
    \mathrm{d} \boldsymbol{x}_t = \mathbf{f}(\boldsymbol{x}_t, t) \, \mathrm{d} t + g(t) \, \mathrm{d} \mathbf{w}
\end{equation}
where $\mathbf{f}(\cdot, \cdot)$ is the drift coefficient, $g(\cdot)$ is the diffusion coefficient, and $\mathbf{w}$ represents the standard Wiener process. Over time, the data will gradually be perturbed into a Gaussian density $\boldsymbol{x}_T \sim \mathcal{N}(0,{\sigma_T}^2I)$. To generate data from noise, diffusion models reverse this process by denoising, as described by:
\begin{equation}\label{reverse}
\mathrm{d} \boldsymbol{x}_t = \left[\mathbf{f}(\boldsymbol{x}_t, t) - g(t)^2 \nabla_{\boldsymbol{x}_t} \log p(\boldsymbol{x}_t) \right] \, \mathrm{d} t + g(t) \, \mathrm{d} \bar{\mathbf{w}}
\end{equation}
where $\bar{\mathbf{w}}$ denotes the reverse-time Wiener process, and $\nabla_{\boldsymbol{x}_t} \log p(\boldsymbol{x}_t)$ is the score, which represents the gradient of the log-likelihood of the marginal distribution at time $t$. The score can be approximated using a neural network by minimizing the denoising score matching loss \citep{vincent2011connection}. 

\subsection{Diffusion Posterior Sampling (DPS)}
Several methods have been proposed to leverage diffusion models as plug-and-play priors for solving inverse problems. A notable method is diffusion posterior sampling (DPS) \citep{chung2022diffusion}. Specifically, DPS replaces the unconditional score in Eq.~\ref{reverse} with a conditional score:
\begin{equation}
\nabla_{\boldsymbol{x}_t} \log p(\boldsymbol{x}_t|\boldsymbol{y}) = \nabla_{\boldsymbol{x}_t} \log p(\boldsymbol{x}_t) + \nabla_{\boldsymbol{x}_t} \log p(\boldsymbol{y}|\boldsymbol{x}_t)
\end{equation}
This modification aims to enable sampling from the posterior distribution $p(\boldsymbol{x}_0|\boldsymbol{y}) $ given the observation. However, the conditional score involves the intractable integral:
\begin{equation}\label{likehood}
p(\boldsymbol{y}|\boldsymbol{x}_t) = \int p(\boldsymbol{y}|\boldsymbol{x}_0) p(\boldsymbol{x}_0|\boldsymbol{x}_t) \, \mathrm{d}\boldsymbol{x}_0 = \mathbb{E}_{p(\boldsymbol{x}_0|\boldsymbol{x}_t)}[p(\boldsymbol{y}|\boldsymbol{x}_0)]
\end{equation}
This integral is computationally expensive, often requiring multiple clean samples of the full trajectory. To address this, DPS approximates the guidance term as follows:
\begin{equation}
p(\boldsymbol{y}|\boldsymbol{x}_t) \approx p(\boldsymbol{y}|\hat{\boldsymbol{x}}_{0|t}) = -\frac{1}{\sigma^2_y}\nabla_{\boldsymbol{x}_t}\|\boldsymbol{y}-\mathcal{H}(\hat{\boldsymbol{x}}_{0|t})\|^2
\end{equation}
where $\hat{\boldsymbol{x}}_{0|t}=\mathbb{E}[\boldsymbol{x}_0|\boldsymbol{x}_t]$ can be computed via Tweedie's formula with one-step denoising \citep{efron2011tweedie}, and $\sigma^2_y$ is the variance of noise in the observation. By employing this approximation, DPS solves the inverse problem without the need for training or finetuning. However, DPS requires a clear gradient of the observation operator, which limits its applicability in scenarios where such gradients are difficult to compute or unavailable.

\subsection{Inverse Problem as Stochastic Control}
We adopt the task formulation proposed by SCG \cite{huang2024symbolic} and frame the solution to the inverse problem within a stochastic control framework. To guide the denoised trajectory toward a target state that satisfies the observation, we introduce a control signal $\boldsymbol{u}_t$ at each time point $t$ in Eq.~\ref{reverse}. This transforms the inverse problem into a stochastic optimal control problem \citep{pavon1989stochastic}, where the terminal cost and transient cost are expressed as follows:
\begin{equation}\mathcal{C}=\mathbb{E}\left[\Phi(\boldsymbol{x}_0)+\int_0^T\frac{\alpha }{2}\|\boldsymbol{u}_t\|^2|\boldsymbol{x}_T\right]
\end{equation}
where $\Phi(\boldsymbol{x}_0)$ represents the terminal cost, and $\alpha$ is a coefficient associated with the transient cost. In the context of the inverse problem, we define it as $\Phi(\boldsymbol{x}_0) = \|\boldsymbol{y}-\mathcal{H}(\boldsymbol{x}_{0})\|^2$. 
By applying path integral control theory \citep{theodorou2010generalized,williams2017model,zhangpath}, we obtain the following optimal policy:
\begin{equation}\label{path}
 \mathbf{u}_t^* \mathrm{d} t =\frac{\mathbb{E}_{p(\boldsymbol{x}_0|\boldsymbol{x}_t)}\left[\exp{(\frac{-\Phi\left(\boldsymbol{x}_0\right)}{\alpha})} \mathrm{d} \bar{\mathbf{w}} \right]}{\mathbb{E}_{p(\boldsymbol{x}_0|\boldsymbol{x}_t)}\left[\exp{(\frac{-\Phi\left(\boldsymbol{x}_0\right)}{\alpha})}  \right]} 
\end{equation}
In this equation, the sample form $p(\boldsymbol{x}_0|\boldsymbol{x}_t)$ is obtained via the uncontrolled process, i.e., Eq.~\ref{reverse}. As can be seen, Eq.~\ref{path} still faces the same challenge as in Eq.~\ref{likehood}, where computing the expected term requires simulating the entire trajectory to obtain $\boldsymbol{x}_0$. SCG \citep{huang2024symbolic} uses an intuitive approximation to estimate the optimal policy. They first consider the limit where $\alpha \rightarrow 0$, in which case Eq.~\ref{path}  degenerates into selecting the $\mathrm{d} \bar{\mathbf{w}}$ that minimizes $\Phi\left(\boldsymbol{x}_0\right)$. Next, they apply an approach similar to DPS, using one-step denoising results $\Phi\left(\hat{\boldsymbol{x}}_{0|t}\right)$ to estimate $\Phi\left(\boldsymbol{x}_0\right)$. In practice, SCG samples multiple candidates at the diffusion reverse kernel $\boldsymbol{x}_t^1,\cdots, \boldsymbol{x}_t^n \sim p(\boldsymbol{x}_t|\boldsymbol{x}_{t+1}) = \mathcal{N}(\mu_t, \sigma_t^2 I)$ of the DDIM solver \cite{song2020denoising} and heuristically retains the particle whose clean estimation is closest to $\boldsymbol{y}$ as the next state. The other particles are discarded.

\section{Methods}
In this section, we first present an empirical analysis to highlight the inefficiencies of the SCG method. Building on these insights, we introduce the \textbf{\emph{Constrained Particle Seeking (CPS)}} framework, which utilizes all available candidate location information to identify the optimal particle. Finally, we explain how CPS can be combined with a \emph{Restart} strategy to reduce cumulative errors and further enhance the robustness of the solution.

\begin{figure}[!t]
    \centering
    \begin{subfigure}[b]{0.8\columnwidth}
        \includegraphics[width=\columnwidth]{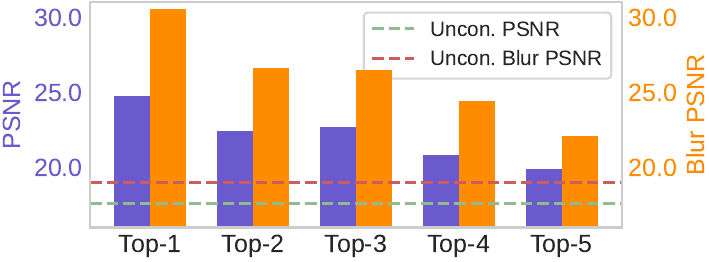}
        \label{fig:image1}
    \end{subfigure}
    \begin{subfigure}[b]{0.8\columnwidth}
        \hspace{0.1em}\includegraphics[width=\columnwidth]{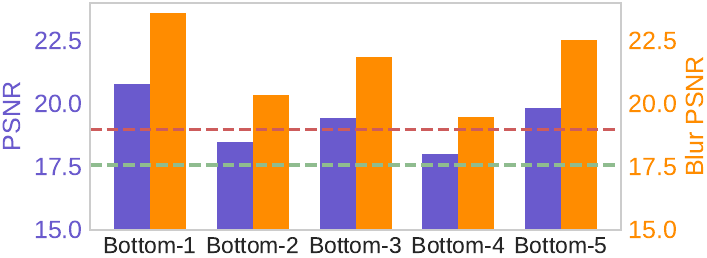}
        \label{fig:image2}
    \end{subfigure}
    \caption{Empirical results of black hole imaging for different particle selections in SCG. The top shows the results obtained by selecting the first few particles in the ranking, while the bottom demonstrates the effect of selecting the last few particles and reversing the sign of their noise.}
    \label{fig:combined}
\end{figure}

\subsection{Leveraging Information from All Particles}\label{method1}
While conceptually simple, methods like SCG \citep{huang2024symbolic} suffer from suboptimal particle efficiency by discarding all but one candidate during the reverse denoising process. We hypothesize that a significant amount of valuable information is lost in this process, as most particles contain insights that can aid in guiding the denoising trajectory. To test this hypothesis, we conduct an empirical study using the black hole imaging inverse problem \citep{zhenginversebench}. 

Our results, depicted in Figure~\ref{fig:combined} (top), provide strong evidence for this hypothesis. Not only does the top-1 particle effectively guide the denoising process towards the observed signal, but the top-2 and top-3 particles also contribute significantly, each outperforming unconditioned sampling. Furthermore, we find a surprising result: even the five worst-performing particles (bottom-1 to bottom-5), when their noise signs are reversed, can still steer the denoising process in the correct direction, as illustrated in Figure~\ref{fig:combined} (bottom). These findings strongly suggest that information useful for guiding the diffusion process is not confined to a single particle but is distributed across the entire ensemble. Inspired by these empirical observations, we propose a novel method to systematically leverage information from all candidates.

\subsection{Constrained Particle Seeking (CPS)}
The SCG approach can be understood as first sampling from the unconditional transition kernel $p(\boldsymbol{x}_t|\boldsymbol{x}_{t+1})$ and then passively selecting the most promising candidates. In contrast, we take a more active approach by seeking promising particles while incorporating additional constraints that align with high-density regions of the unconditional prior. We reformulate this idea as the following constrained optimization problem:
\begin{equation}\label{cons}
    \boldsymbol{x}_t^*= \arg\min_{\boldsymbol{x}_t} \Phi\left(\hat{\boldsymbol{x}}_{0|t}\right)\quad \text{s.t.} \ \boldsymbol{x}_t \sim p(\boldsymbol{x}_t|\boldsymbol{x}_{t+1})
\end{equation}
This optimization framework offers greater flexibility, allowing for more effective particle seeking, as it is no longer confined to selecting particles from the sampled set alone.

Solving this optimization problem presents two main challenges. First, we must address how to minimize the objective function, particularly when dealing with black-box observation processes. Second, we need to implement the constraint of the unconditional kernel in practice. In the following, we will outline our solutions to these challenges.

\begin{figure}[!t]
 \includegraphics[width=1.0\columnwidth]{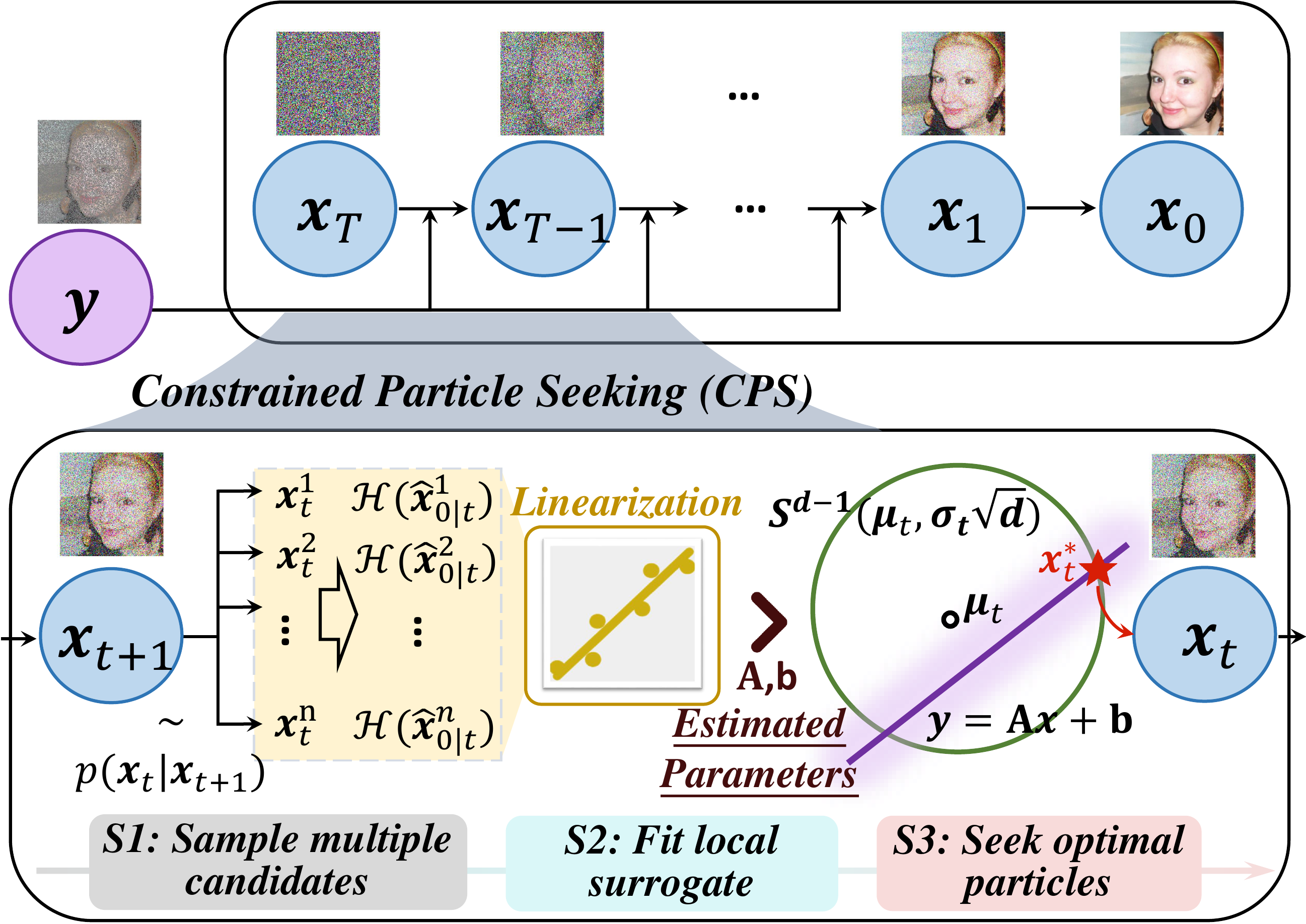}
    \caption{Overview of Constrained Particle Seeking (CPS) for inverse problem solving. At each time step, the process involves three steps. \emph{Step 1:} Sample multiple candidate particles; \emph{Step 2:} Fit the forward process surrogate using these samples; \emph{Step 3:} Use the surrogate to determine the optimal particle on the hypersphere, as the next state.}
    \label{method}
\end{figure}

\noindent \textbf{Surrogate of observation process.} Optimizing the terminal cost typically requires estimating its gradient with respect to $\boldsymbol{x}_t$, which can be derived as follows:
\begin{equation}
    \nabla_{\boldsymbol{x}_t}\Phi\left(\hat{\boldsymbol{x}}_{0|t}\right) = \left(\frac{\partial \mathcal{H}( \hat{\boldsymbol{x}}_{0|t})}{\partial \boldsymbol{x}_t}\right)^\top \left( \mathcal{H}( \hat{\boldsymbol{x}}_{0|t})-\boldsymbol{y}\right)
\end{equation}
However, computing the term $\frac{\partial \mathcal{H}( \hat{\boldsymbol{x}}_{0|t})}{\partial \boldsymbol{x}_t}$ is challenging, as we do not have direct access to the gradient $\mathcal{H}(\cdot)$. To address this, we apply statistical linearization \citep{roberts2003random} to estimate a local surrogate around the region defined by $p(\boldsymbol{x}_t|\boldsymbol{x}_{t+1})$. Specifically, we first sample $n$ particles $\boldsymbol{x}_t^1,\cdots, \boldsymbol{x}_t^n \sim p(\boldsymbol{x}_t|\boldsymbol{x}_{t+1})$,  as done in SCG, and estimate the corresponding values $\mathcal{H}\left(\hat{\boldsymbol{x}}_{0|t}^1\right),\cdots,\mathcal{H}\left(\hat{\boldsymbol{x}}_{0|t}^n\right)$. We then fit these values to the best-fitting linear model with parameters $(\mathbf{A}, \mathbf{b})$. This leads to the following optimization problem:
\begin{equation}
    \min_{\mathbf{A}, \mathbf{b}} \mathbb{E}_{\boldsymbol{x}_t\sim p(\boldsymbol{x}_t|\boldsymbol{x}_{t+1})}\|\mathcal{H}\left(\hat{\boldsymbol{x}}_{0|t}\right)-(\mathbf{A}\boldsymbol{x}_t+\mathbf{b})\|^2 
\end{equation}
The closed-form solution to this problem is given by:
\begin{equation}\label{A,b}
\begin{aligned}
    &\mathbf{A}= \text{Cov}(\mathcal{H}\left(\hat{\boldsymbol{x}}_{0|t}\right),\boldsymbol{x}_t) \text{Cov}(\boldsymbol{x}_t, \boldsymbol{x}_t)^{-1}\\ &\mathbf{b}  = \mathbb{E}[\mathcal{H}\left(\hat{\boldsymbol{x}}_{0|t}\right)]-\mathbf{A}\mathbb{E}[\boldsymbol{x}_t]
\end{aligned}
\end{equation}
Given that the transition kernel of the diffusion model is an isotropic Gaussian density, we represent its mean and variance as $\mathcal{\mu}_t$ and $\sigma_t^2$, respectively.  This enables us to further simplify the expressions. Next, we perform empirical estimation  using the sampled particles $\boldsymbol{x}_t^1,\cdots, \boldsymbol{x}_t^n$, yielding the following expression for $\mathbf{A}$ and $\mathbf{b}$,
\begin{equation}\label{part_a}
\begin{aligned}
        &\mathbf{A}=\frac{1}{n \sigma_t^2}\sum_{i=1}^{n}(\mathcal{H}\left(\hat{\boldsymbol{x}}_{0|t}^i\right)-\bar{\mathcal{H}})(\boldsymbol{x}_t^i-\mathcal{\mu}_t)^\top\\
        &\mathbf{b}=\bar{\mathcal{H}}-\mathbf{A}\mu_t
\end{aligned}
\end{equation}
where $\bar{\mathcal{H}}$ denotes $\frac{1}{n}\sum_{i=1}^n \mathcal{H}\left(\hat{\boldsymbol{x}}_{0|t}^i\right)$ for simplicity. At this point, we can use the surrogate model to replace the non-differentiable part of the terminal cost. Furthermore, we observe that $\mathbf{A}$ in Eq.~\ref{part_a} is effectively a weighted linear combination of the sampled particles, integrating all candidate information. This enables us to select particles that outperform any individual sample, thus facilitating a more effective identification of the next state.


\noindent  \textbf{Constraint of transition kernel.} In the previous section, we introduced the use of a linear surrogate to estimate terminal costs, enabling us to optimize the objective. Eq.~\ref{cons} also imposes the condition that the optimization variables be sampled from an unconditional distribution, ensuring that the surrogate remains within a credible range and that the model does not deviate from the prior of the base model during inference. 

The transition kernel $p(\boldsymbol{x}_t|\boldsymbol{x}_{t+1})$ is a Gaussian distribution with a tractable log-likelihood function and it tends to push $\boldsymbol{x}_t$ toward the mean. As shown by \citet{menon2020pulse}, the norm of samples from a $d$-dimensional Gaussian distribution follows a $\chi^d$ distribution. For the high-dimensional spaces common in diffusion models, the mass of the $\chi^d$ distribution is highly concentrated around the value $\sigma_t\sqrt{d}$, forming a hypersphere $S^{d-1}(\mu_t,\sigma_t\sqrt{d})$ \citep{samuel2023norm}. Diffusion models generally prefer input data that follows a norm-based distribution aligned with this characteristic, rather than being centered around the mean. Therefore, we aim to keep $\boldsymbol{x}_t$ within the high-mass region of $p(\boldsymbol{x}_t|\boldsymbol{x}_{t+1})$. With this in mind, the original optimization problem (Eq.~\ref{cons}) can be interpreted as follows, using the surrogate model:
\begin{equation}
    \min_{\boldsymbol{x}_t}\|\boldsymbol{y}-(\mathbf{A}\boldsymbol{x}_t+\mathbf{b})\|^2 \quad \text{s.t.} \ \boldsymbol{x}_t \in S^{d-1}(\mu_t,\sigma_t\sqrt{d})
\end{equation}
The aforementioned constrained optimization problem can be solved using the Lagrange multiplier method \citep{bertsekas2014constrained}, leading to the following system of equations for the optimal solution:
\begin{equation}
    \begin{cases}
        \mathbf{A}^\top(\mathbf{A}\boldsymbol{x}_t+\mathbf{b}-\boldsymbol{y})
    +\lambda(\boldsymbol{x}_t-\mu_t)=0\\
    \|\boldsymbol{x}_t-\mu_t\|^2-\sigma_t^2d=0
    \end{cases}
\end{equation}
where $\lambda$ is the Lagrange multiplier. Solving this system directly involves matrix inversion and root finding, which becomes challenging in high-dimensional spaces. We note that the $\sigma_t$ of the reverse kernel is typically small,  leading to constraints that dominate when seeking the optimal solution.  As a result, the multiplier $\lambda$ is expected to be sufficiently large. Under these conditions, we can analytically derive the asymptotic form of the optimal solution:
\begin{equation}
    \boldsymbol{x}_t^*\approx \mu_t + \sigma_t\sqrt{d}\frac{\mathbf{A}^\top(\boldsymbol{y}-\bar{\mathcal{H}})}{\|\mathbf{A}^\top(\boldsymbol{y}-\bar{\mathcal{H}})\|}
\end{equation}
A detailed derivation can be found in Appendix A. Ultimately, the solution $\boldsymbol{x}_t^*$ can be viewed as the optimal particle synthesized by combining all candidate information. In comparison to SCG, which merely selects the particle with the minimum terminal cost, our method is more principled and efficient. Figure 3 provides an overview of our approach.

\begin{figure}[!t]
     \centering
\includegraphics[width=0.8\columnwidth]{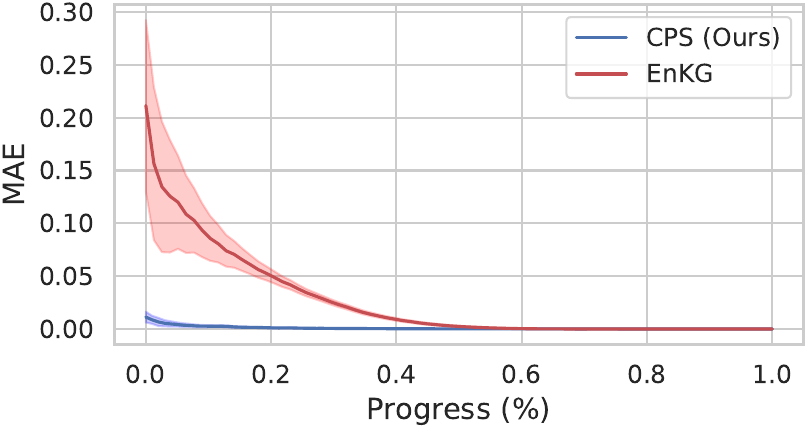}
    \caption{Evolution of Jensen gap of CPS and EnKG for 4$\times$ super-resolution on the FFHQ dataset.}
    \label{enkg}
\end{figure}

Note that EnKG \citep{zhengensemble} employs a similar linearization strategy. However, EnKG performs linearization over the entire global space, which can lead to inaccurate approximations and difficulties when inverting the covariance matrix (Eq.\ref{A,b}). In contrast, our approach utilizes local linearization around $p(\boldsymbol{x}_t|\boldsymbol{x}_{t+1})$. In Figure~\ref{enkg}, we compute the mean absolute error (MAE) of the Jensen gap \cite{gao2017bounds} for the function $\mathcal{H}\left(\mathbb{E}[\boldsymbol{x}_0|\mathbf{\cdot}]\right)$ using sampled particles to assess the degree of deviation from linear behavior. Our results demonstrate that EnKG exhibits significant errors in the early stages of generation, whereas CPS consistently maintains small errors. Additionally, CPS seeks one optimal particle at each step by synthesizing candidates, without explicitly maintaining the entire particle set as in EnKG. Our ablation in the experiment section reveals that CPS are significantly more efficient than those generated by the latter.

\subsection{Improving Robustness with Restart} In practice, we observe that different initial noise conditions can lead to varied outcomes in image restoration tasks. When the initial noise is suboptimal, large adjustments are needed to correct for the global mismatch. On top of that, the one-step estimation from Tweedie's formula can introduce approximation errors that hurt performance.

To address these challenges, we incorporate the recently proposed Restart strategy \citep{xu2023restart, lugmayr2023inpainting}. This method provides a powerful way to mitigate the accumulation of errors throughout the sampling process. The procedure is a simple, two-step iteration: at time step $t$, the sample is re-noised according to Eq.~\ref{forward}, then returned to the previous timestep $t+1$. This re-noised sample is subsequently reused with CPS for guided denoising. Such a stepwise approach allows for the gradual correction of cumulative errors, thereby enhancing the robustness of CPS and ensuring more reliable solutions. The full sampling process, including the Restart strategy, is detailed in Algorithm~\ref{alg:scg_ddpm}.

\begin{algorithm}[!t]
\caption{Sampling procedure for CPS}
\label{alg:scg_ddpm}
\begin{algorithmic}[1]
\Require Observation $\boldsymbol{y}$, initial noise $\boldsymbol{x}_T \sim \mathcal{N}(0, I)$, pretrained diffusion model $p(\boldsymbol{x}_{t}|\boldsymbol{x}_{t+1}) = \mathcal{N}(\mu_t, \sigma_t I)$, data dimension $d$, and number of restart $N_r$
\For{$t = T-1,\cdots,0$}
    \For{$r = N_r,\cdots,1$}
    \Statex \quad \quad \quad \textcolor{gray}{$\triangleright$  \emph{Sample from reverse kernel}}
    \State $\boldsymbol{x}_{t}^1,\cdots,\boldsymbol{x}_{t}^n \sim p(\boldsymbol{x}_{t}|\boldsymbol{x}_{t+1})$
    \Statex \quad \quad \quad \textcolor{gray}{$\triangleright$   \emph{Fit the forward process surrogate }}
    \State 
    $\mathbf{A}\leftarrow\frac{1}{n \sigma_t^2}\sum_{i=1}^{n}(\mathcal{H}\left(\hat{\boldsymbol{x}}_{0|t}^i\right)-\bar{\mathcal{H}})(\boldsymbol{x}_t^i-\mathcal{\mu}_t)^\top$
    \Statex \quad \quad \quad \textcolor{gray}{$\triangleright$   \emph{Solve optimal particle as next state}}
    \State $\boldsymbol{x}_t \leftarrow \mu_t + \sigma_t\sqrt{d}\frac{\mathbf{A}^\top(\boldsymbol{y}-\bar{\mathcal{H}})}{\|\mathbf{A}^\top(\boldsymbol{y}-\bar{\mathcal{H}})\|}$
    \If{$r>1$}
        \Statex \quad \quad \quad \quad \ \ \textcolor{gray}{$\triangleright$   \emph{Re-noise for restart}}
        \State $\boldsymbol{x}_{t+1} \leftarrow p(\boldsymbol{x}_{t+1}|\boldsymbol{x}_{t})$
    \EndIf
    \EndFor
\EndFor
\Ensure $\boldsymbol{x}_0$
\end{algorithmic}
\end{algorithm}

\begin{table*}[t]
\centering
  \resizebox{1\textwidth}{!}{
\begin{tabular}{clcccccccccccc}
   \toprule
&  &  \multicolumn{3}{c}{\bf Inpaint (95\%)}& \multicolumn{3}{c}{ \bf Super-resolution ($\times$4)}& \multicolumn{3}{c}{ \bf Deblur (gauss)} & \multicolumn{3}{c}{ \bf JPEG (QF=5)}\\
   \cmidrule(r){3-5}\cmidrule(r){6-8}\cmidrule(r){9-11}\cmidrule(r){12-14}
   & \bf Method& PSNR $\uparrow$ &SSIM $\uparrow$ & LPIPS $\downarrow$ & PSNR $\uparrow$ &SSIM $\uparrow$ & LPIPS $\downarrow$ & PSNR $\uparrow$ &SSIM $\uparrow$ & LPIPS $\downarrow$& PSNR $\uparrow$ &SSIM $\uparrow$ & LPIPS $\downarrow$\\
\midrule
& DDRM &21.65&0.693&0.270&\cellcolor{mysec!50}28.02&0.828&0.213& 24.69&0.753&0.187&-&-&-\\
& DPS & 22.49&0.701&0.236&27.65&0.822&\cellcolor{mysec!50}0.120&25.59&0.790&\cellcolor{mythird!50}0.136&-&-&-\\
Gradient &$\Pi$GDM&\cellcolor{mythird!50}23.13&\cellcolor{mythird!50}0.774&0.231&27.25&0.829&0.135&26.76&0.812&0.141&18.25&0.593&0.304 \\
 -based& RED-diff&\cellcolor{mysec!50}24.43& \cellcolor{mytop!50}\bf 0.807&\cellcolor{mysec!50}0.222& 26.85&\cellcolor{mysec!50} 0.835&0.191&\cellcolor{mysec!50}28.01& \cellcolor{mytop!50}\bf 0.853&0.164 &-&-&-\\
 & DAPS & 22.71& 0.754&\cellcolor{mythird!50}0.226& \cellcolor{mytop!50}\bf 28.53& \cellcolor{mytop!50}\bf 0.840&\cellcolor{mythird!50}0.132&\cellcolor{mytop!50} \bf 28.70& \cellcolor{mysec!50}0.843&\cellcolor{mysec!50}0.133&-&-&-\\
 \midrule
  & SCG &6.15& 0.321& 0.807& 6.16&0.319&0.805& 7.01&0.322&0.792&6.17&0.362&0.825\\
 Gradient & DPG&19.83&0.651&0.299& 24.81&0.758&0.187&24.11&0.740&0.196&\cellcolor{mysec!50}22.52 &\cellcolor{mysec!50}0.734&\cellcolor{mysec!50}0.246\\
 -free&EnKG&22.64& 0.767& 0.286&21.72&0.737&0.293& 20.84& 0.731&0.334&\cellcolor{mythird!50}20.89&\cellcolor{mythird!50}0.704&\cellcolor{mythird!50}0.257\\
 & \bf CPS (Ours)&\cellcolor{mytop!50} \bf 24.90&\cellcolor{mysec!50}0.794& \cellcolor{mytop!50}\bf 0.161&\cellcolor{mythird!50}27.70&\cellcolor{mythird!50}0.834& \cellcolor{mytop!50}\bf 0.115&\cellcolor{mythird!50}27.21&\cellcolor{mythird!50}0.822& \cellcolor{mytop!50}\bf 0.128 &\cellcolor{mytop!50} \bf 23.23&\cellcolor{mytop!50} \bf 0.784&\cellcolor{mytop!50} \bf 0.223\\
   \bottomrule
\end{tabular}}
\caption{Quantitative results of different methods on FFHQ, including three linear and a non-differentiable inverse problems. Best results are highlighted as \colorbox{mytop!50}{\textbf{first}}, \colorbox{mysec!50}{second} and \colorbox{mythird!50}{third}. We mark "-" for those not applicable.}\label{tab:image}
\end{table*}

\begin{figure}[!t]
     \centering
\includegraphics[width=1\columnwidth]{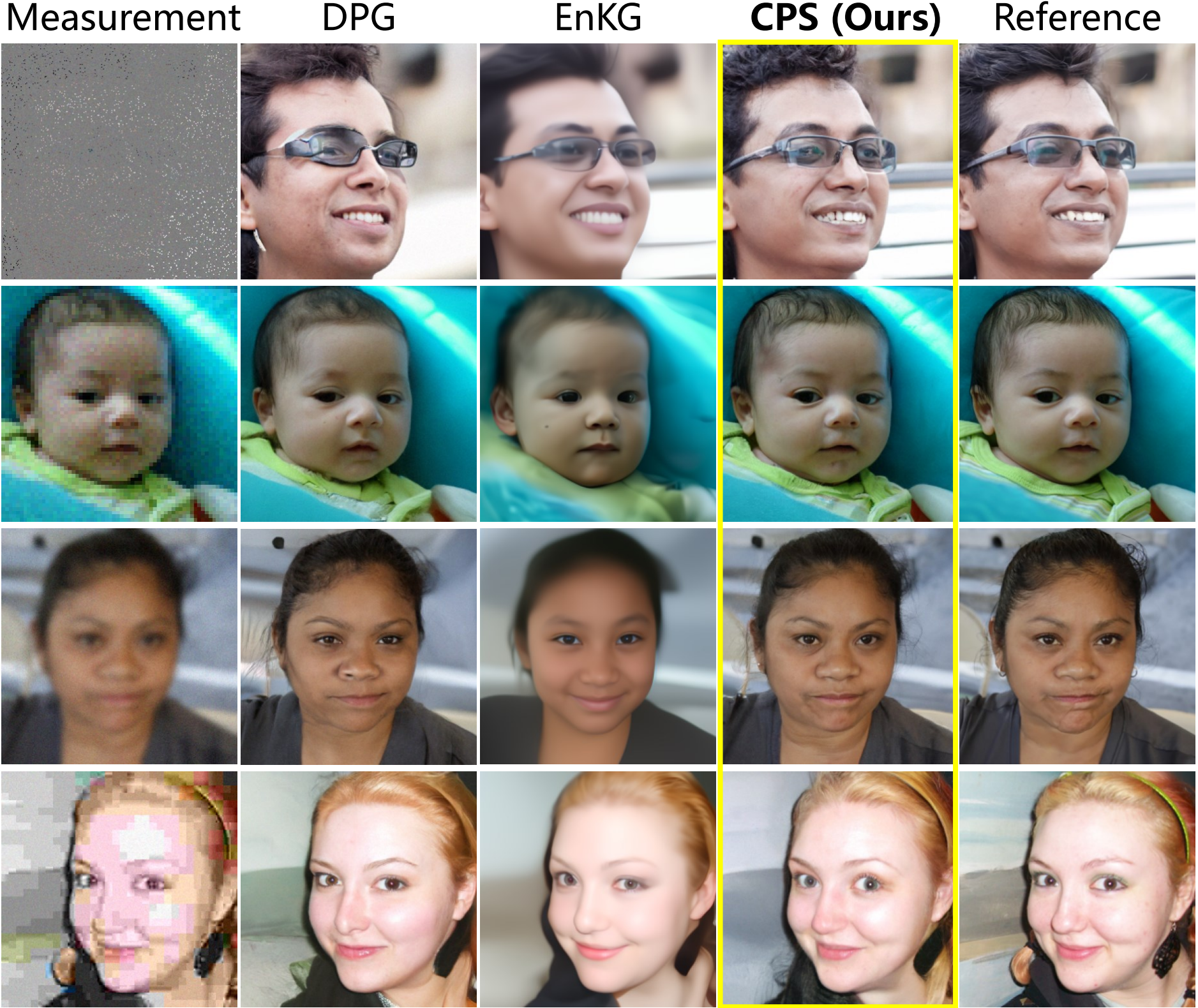}
    \caption{Qualitative results of image inverse problems on FFHQ: from top to bottom, inpainting, super-resolution, deblurring, and JPEG restoration.}
    \label{image_fig}
\end{figure}

\section{Experiments}
To validate the effectiveness of our method, we conduct experiments on image inverse problems and two scientific inverse problems. Our baseline includes several recently proposed gradient-free methods for solving inverse problems, such as SCG \cite{huang2024symbolic}, DPG \cite{tang2024solving}, and EnKG \cite{zhengensemble}. All gradient-free methods are evaluated using a comparable number of forward passes to ensure a fair comparison. Additionally, we incorporate advanced gradient-based methods for comparison in the linear image problem, including DDRM \cite{kawar2022denoising}, DPS \cite{chung2022diffusion}, $\Pi$GDM \cite{song2023pseudoinverse}, RED-diff \cite{mardani2023variational}, and DAPS \cite{zhang2025improving}. The formulations of the inverse problems tested are detailed in Appendix B.1. Further implementation details can be found in Appendix B.2.

\subsection{Image Inverse Problem}
\textbf{Experimental Setup:} We begin by evaluating our method on the image inverse problem. We sample 100 images from the FFHQ dataset \citep{karras2019style} for validation and use the checkpoint provided by \citet{chung2022diffusion}. All images were resized to 256 × 256 pixels, and their quality was assessed using three standard image restoration metrics: PSNR, SSIM, and LPIPS \cite{zhang2018unreasonable}.

We focus on three linear inverse problems: 95\% missing random mask completion, 4$\times$ super-resolution, and deblurring with a Gaussian blur kernel of standard deviation 3 (61 × 61). To assess the practicality of gradient-free methods, we also include JPEG restoration tasks \cite{song2023pseudoinverse}, where the forward model is non-differentiable. For the JPEG restoration, we consider the quality factor (QF) which is 5. All measurements included additive Gaussian noise with $\sigma_{y}=0.05$.

\noindent \textbf{Result:} Table~\ref{tab:image} and Figure~\ref{image_fig} present the quantitative results and visualizations, respectively. Our findings show that SCG fails to handle image inverse problems effectively. In contrast, CPS significantly outperforms gradient-free methods and competes well with gradient-based methods. To the best of our knowledge, $\Pi$GDM is the only method capable of performing JPEG restoration through pseudo-inverse construction \cite{song2023pseudoinverse}. However, we observe that $\Pi$GDM struggles in noisy scenes, with its performance being notably poor. This emphasizes the importance of further developing gradient-free inversion methods.

\begin{table}[ht]
    \centering
     \resizebox{0.47\textwidth}{!}{
    \begin{tabular}{lcccccc}
    \toprule
        & \multicolumn{2}{c}{\bf 100\%}& \multicolumn{2}{c}{\bf 10\%}& \multicolumn{2}{c}{\bf 3\%}\\
        \cmidrule(r){2-3}\cmidrule(r){4-5}\cmidrule(r){6-7}
         \bf Method&PSNR $\uparrow$ &BPSNR $\uparrow$& PSNR $\uparrow$ &BPSNR $\uparrow$& PSNR $\uparrow$ &BPSNR $\uparrow$\\
         \midrule
         SCG&\cellcolor{mysec!50}25.45&\cellcolor{mysec!50}30.56&\cellcolor{mysec!50}24.97&\cellcolor{mysec!50}31.45&\cellcolor{mysec!50}24.65&\cellcolor{mysec!50}31.28 \\
         DPG&14.51&16.75&13.57&16.35&12.16&13.71\\
         EnKG&23.03&25.67&22.77&25.18&21.41&25.14\\
         \bf CPS (Ours) &\cellcolor{mytop!50} \bf25.74&\cellcolor{mytop!50} \bf31.84&\cellcolor{mytop!50} \bf25.34&\cellcolor{mytop!50} \bf31.54&\cellcolor{mytop!50} \bf25.03&\cellcolor{mytop!50} \bf31.81\\
         \bottomrule
    \end{tabular}}
    \caption{ Comparisons of gradient-free methods on black hole imaging problem, including three observation time ratios: 100\%, 10\% and 3\%. }
    \label{tab:blackhole}
\end{table}

\begin{figure}[!t]
     \centering
\includegraphics[width=1\columnwidth]{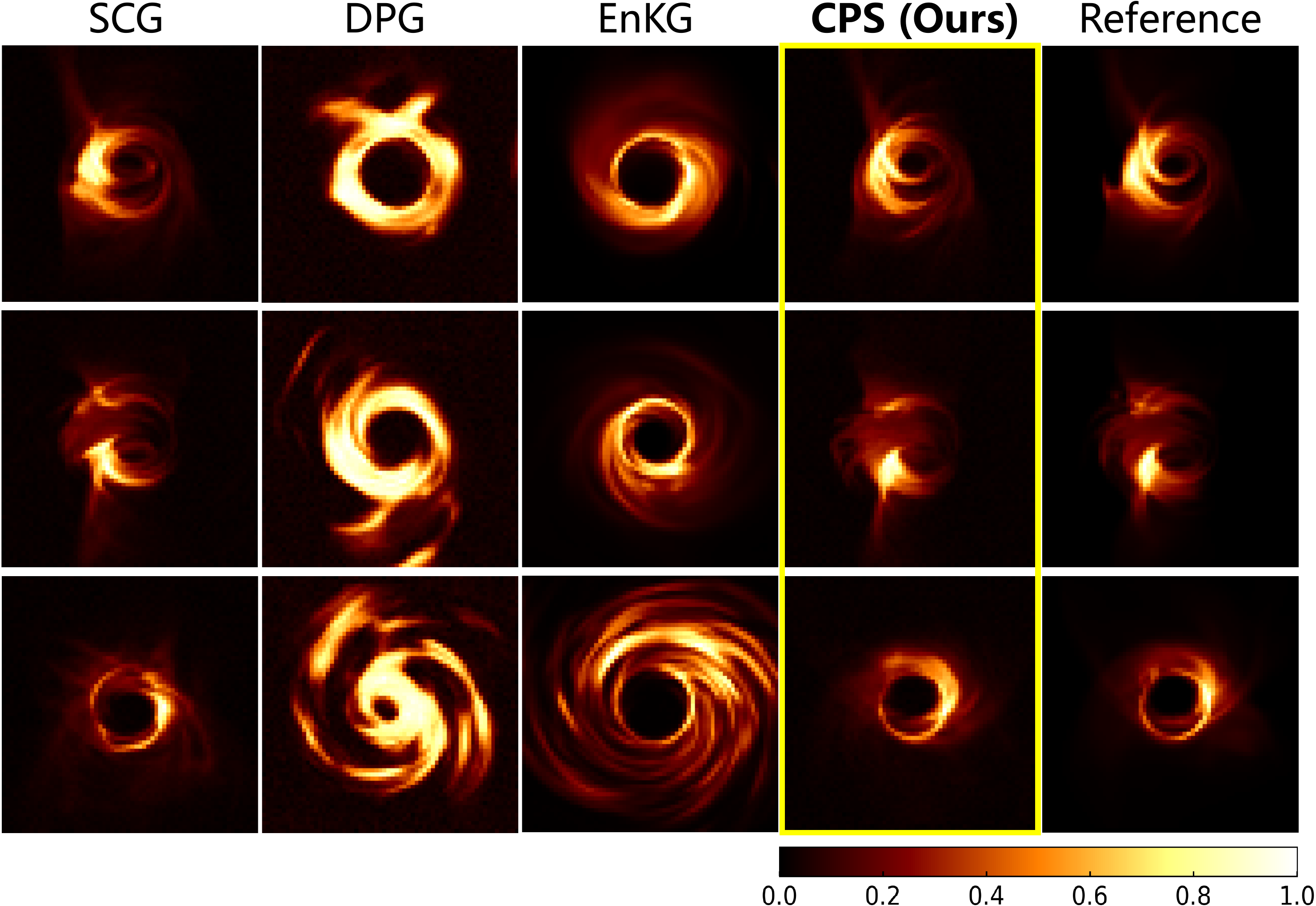}
    \caption{Qualitative results of black hole imaging.}
    \label{bh_fig}
\end{figure}

\begin{figure*}[!t]
    \centering
    \subfloat[4× image super-resolution]{\includegraphics[width=0.335\textwidth]{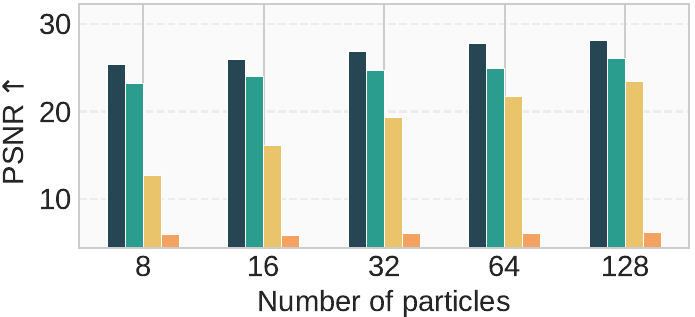}\label{fig:sub1}}
    \hfill
    \subfloat[Black hole imaging]{\includegraphics[width=0.335\textwidth]{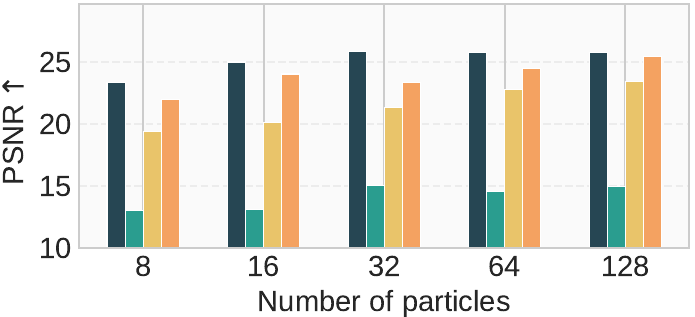}\label{fig:sub2}}
    \hfill
    \subfloat[Fluid Data Assimilation]{\includegraphics[width=0.325\textwidth]{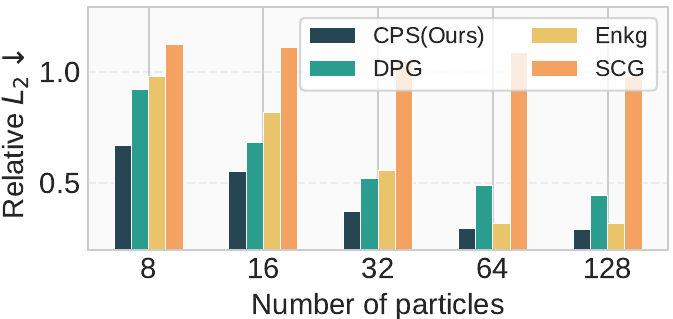}\label{fig:sub3}}
    \caption{Comparison of particle efficiency of gradient-free method across three tasks.}
    \label{fig:three_subfigs}
\end{figure*}

\subsection{Black Hole Imaging}
\textbf{Experimental Setup:} Black hole imaging presents a highly nonlinear and ill-posed forward model due to sparse observations. Measurements are obtained via Very Long Baseline Interferometry (VLBI), where each telescope pair provides visibility data at specific times. To reduce errors from atmospheric turbulence and thermal noise, multiple visibility data points are typically combined into \emph{closure phases} and \emph{log closure amplitudes} \cite{blackburn2020closure}. In addition to these constraints, the \emph{total flux} of the black hole is often incorporated into the likelihood function. For evaluation, we assume black-box access to the forward model. We employ a pretraining diffusion model from \citet{zhenginversebench} and evaluate it on 100 resized 64$\times$64 images from the test set. We consider the settings of three observation time ratios: 100\%, 10\% and 3\%.
The reconstruction quality is measured by calculating the PSNR between the reconstructed image and the ground truth, along with the Blur PSNR (BPSNR), which is smoothed using a filter to assess low-frequency reconstruction performance.

\noindent \textbf{Result:} Figure~\ref{bh_fig} compares black hole images reconstructed using gradient-free method. The CPS results are visually more consistent with the ground truth. Table~\ref{tab:blackhole} presents the quantitative comparison, with CPS achieving the best performance, followed closely by SCG. In contrast, DPG performs poorly when faced with such highly nonlinear problems.

\begin{table}[ht]
    \centering
     \resizebox{0.4\textwidth}{!}{
    \begin{tabular}{lccc}
    \toprule
        & \multicolumn{1}{c}{\bf $\times$2}& \multicolumn{1}{c}{\bf $\times$4}& \multicolumn{1}{c}{\bf $\times$8}\\
        \cmidrule(r){2-4}
         \bf Method&Relative $L_2$ $\downarrow$ &Relative $L_2$ $\downarrow$ &Relative $L_2$ $\downarrow$ \\
         \midrule
         SCG&1.087&1.095&1.123\\
         DPG&0.491&0.592&0.836\\
         EnKG&\cellcolor{mysec!50}0.320&\cellcolor{mysec!50}0.528&\cellcolor{mysec!50}0.821\\
         \bf CPS (Ours) &\cellcolor{mytop!50} \bf0.295&\cellcolor{mytop!50} \bf0.469&\cellcolor{mytop!50} \bf0.684\\
         \bottomrule
    \end{tabular}}
    \caption{ Comparisons of gradient-free methods on fluid aata assimilation, including three types of sparse observations: 2$\times$, 4$\times$, and 8$\times$ downscaling. }
    \label{tab:ns}
\end{table}

\begin{figure}[htbp]
     \centering
\includegraphics[width=1\columnwidth]{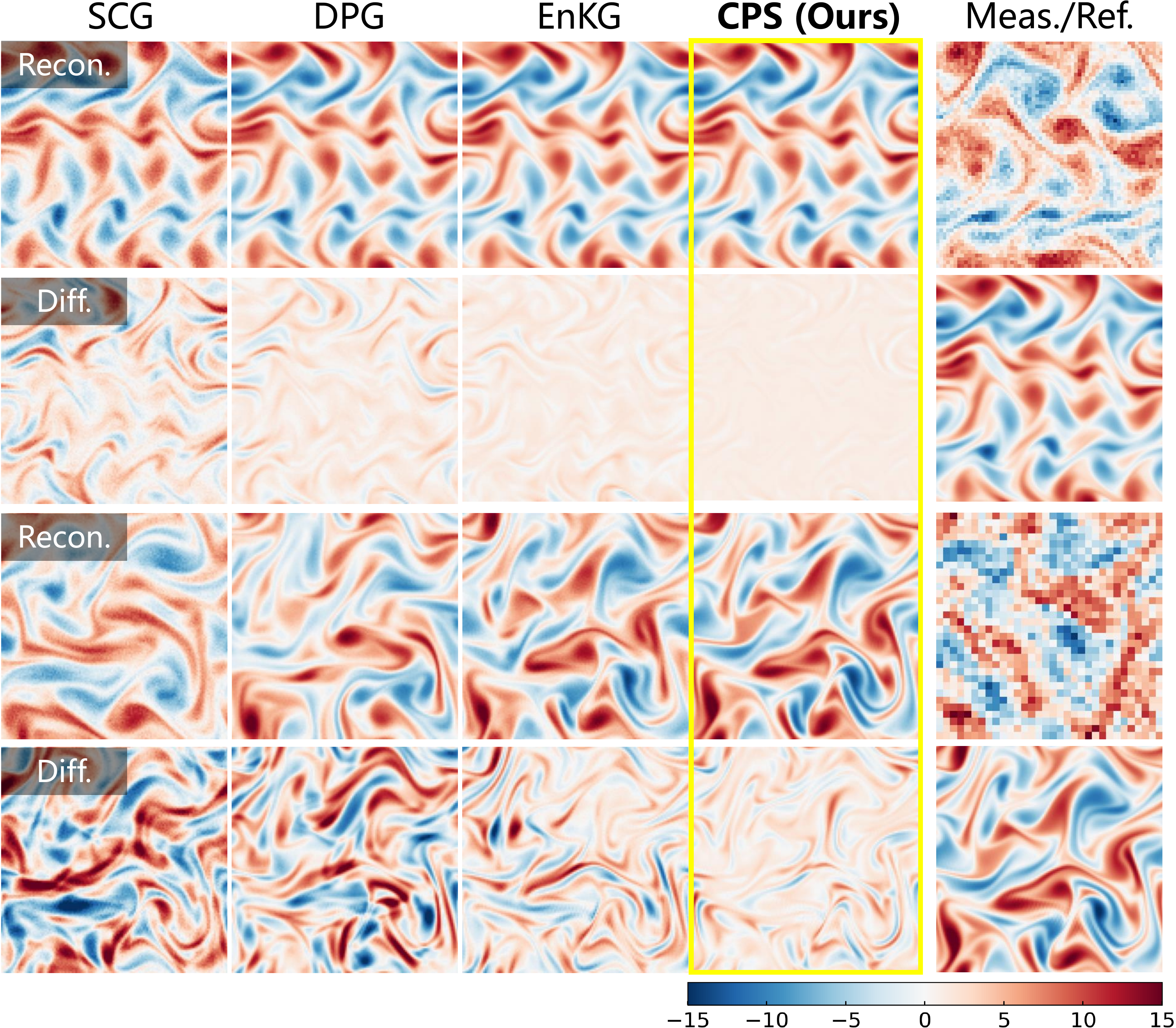}
    \caption{Qualitative results of recovering initial vorticity fields. Odd rows show the reconstruction, even rows show the difference from the reference.}
    \label{fluid_fig}
\end{figure}

\subsection{Fluid Data Assimilation} 
\noindent \textbf{Experimental Setup:} Fluid dynamics are typically modeled by the Navier-Stokes (NS) equations, a set of nonlinear PDEs \cite{chandler2013invariant}. In this experiment, we perform fluid assimilation on sparse and noisy vorticity measurements at $T=1$, with the goal of recovering the 2D initial vorticity field. Given that PDE numerical solvers may require thousands of discrete time steps, obtaining accurate numerical derivatives through automatic differentiation presents significant challenges. This makes the problem an ideal candidate for validating gradient-free methods. We use the pretrained checkpoint provided by \citet{zhenginversebench} and 10 vorticity fields with size 128$\times$128 for testing. The result is evaluated using the relative $L_2$ error. We consider three settings to simulate sparse measurements, including 2$\times$, 4$\times$, and 8$\times$downscaling. All measurements included noise with $\sigma_y=2.0$. 

\noindent \textbf{Result:} Figure~\ref{fluid_fig} presents both a simple example and a challenging visualization result. In the simple example, all methods are able to restore the overall structure effectively. However, in more difficult scenarios, CPS significantly outperforms the baseline method by recovering the initial vorticity field pattern from sparse and noisy measurements. SCG and DPG often fail under these conditions. EnKG performs second best, but with noticeable errors in some local areas. The quantitative results in Table~\ref{tab:ns} further support our findings.


\subsection{Ablation Study}
\textbf{Particle Efficiency Comparison:} Gradient-free methods typically require sampling multiple particles. In this study, we evaluate the particle efficiency of these methods using particle numbers of 8, 16, 32, 64, and 128. As shown in Figure~\ref{fig:three_subfigs}, CPS demonstrates significantly higher particle efficiency compared to the baseline method, maintaining stability and competitiveness across various tasks. Even with just 8 particles, CPS performs well. In contrast, EnKG struggles with small particle sets and requires a larger number of particles to function effectively. DPG faces difficulties with highly ill-posed black hole imaging, while SCG fails to address image-related problems.

\noindent \textbf{Effectiveness of Restart:} We investigate the impact of the Restart strategy in CPS, using different random seeds to initialize the noise. As shown in Appendix C, the Restart strategy not only improves overall performance but also enhances the stability and robustness of our algorithm. It effectively addresses the issue of sampling from poor initial noise, preventing failures during the inversion process.

\section{Conclusion}
In this work, we introduced \textbf{\emph{CPS}}, a novel gradient-free method for solving inverse problems with diffusion models. CPS uses information from all candidate particles and applies constraints based on high-density regions of the prior, enables efficient particle seeking. Our experiments show that CPS outperforms other gradient-free methods and achieves results similar to gradient-based approaches. These results demonstrate CPS's potential as a reliable solution when gradients of the forward process are not available.

\section{Acknowledgments}
This work was supported by the National Natural Science Foundation of China under Grant 62405014, 62325101, and 62031001.
\bibliography{aaai2026}

\clearpage
\section{Appendix A: Derivations}

\subsection{A1: Derivation of Eq. 16 of Main Text}
\textbf{Lagrange Multiplier Method:} For the optimization problem on the sphere, as presented in Eq. 14 of the main text, we aim to solve:
\begin{equation}
\min _{\boldsymbol{x}_t}\left\|\boldsymbol{y}-\left(\mathbf{A} \boldsymbol{x}_t+\mathbf{b}\right)\right\|^2 \quad \text { s.t. } \|\boldsymbol{x}_t-\mu_t\|^2 =\sigma_t^2 d
\end{equation}
We can solve this problem by relaxing it with the Lagrange multiplier method. The Lagrangian function is:
\begin{equation}
    L = \left\|\boldsymbol{y}-\left(\mathbf{A} \boldsymbol{x}_t+\mathbf{b}\right)\right\|^2 +\lambda (\|\boldsymbol{x}_t-\mu_t\|^2 -\sigma_t^2 d)
\end{equation}
By setting the partial derivatives with respect to the variables $\boldsymbol{x}_t$ and the multiplier $\lambda$ to zero, we obtain the following system of equations for the optimal solution:
\begin{equation}
\left\{\begin{array}{l}
\mathbf{A}^{\top}\left(\mathbf{A} \boldsymbol{x}_t+\mathbf{b}-\boldsymbol{y}\right)+\lambda\left(\boldsymbol{x}_t-\mu_t\right)=0 \\
\left\|\boldsymbol{x}_t-\mu_t\right\|^2-\sigma_t^2 d=0
\end{array}\right.
\end{equation}
This system of equations can be solved to find the optimal candidate particles. However, an analytical solution is generally not available, so we would typically resort to numerical methods. Specifically, the first equation can be rewritten to form a linear system for $\boldsymbol{x}_t$:
\begin{equation}
    (\mathbf{A}^\top\mathbf{A}+\lambda I)\boldsymbol{x}_t=\mathbf{A}^\top(\boldsymbol{y}-\mathbf{b})+\lambda \mu_t
\end{equation}
This system is often solved using iterative methods, such as the conjugate gradient (CG) method \cite{daniel1967conjugate} or GMRES \cite{saad1986gmres}. After an estimate for $\boldsymbol{x}_t(\lambda)$ is obtained, it is substituted into the second equation, which can then be solved for $\lambda$ using a root-finding algorithm such as Newton's method \cite{broyden1965class}:
\begin{equation}\label{KKT}
    \begin{aligned}
        &\left\|\boldsymbol{x}_t(\lambda)-\mu_t\right\|^2-\sigma_t^2 d=0\\
        \text{s.t.} \quad & \boldsymbol{x}_t(\lambda) = (\mathbf{A}^\top\mathbf{A}+\lambda I)^{-1} [\mathbf{A}^\top(\boldsymbol{y}-\mathbf{b})+\lambda \mu_t]
    \end{aligned}
\end{equation}
By repeating these two steps, we can iteratively converge to a solution. However, this iterative approach, which involves both linear system solving and root-finding, can be computationally expensive and numerically unstable. Therefore, in the following, we propose an approximate closed-form solution that circumvents these issues.

\noindent \textbf{Approximate Solution:} We begin by applying a "skillful" transformation to the expression for $\boldsymbol{x}_t(\lambda)$:
\begin{equation}
    \begin{aligned}
        \boldsymbol{x}_t(\lambda) & = (\mathbf{A}^\top\mathbf{A}+\lambda \mathbf{I})^{-1} [\mathbf{A}^\top(\boldsymbol{y}-\mathbf{b}) + \lambda \mu_t]\\
        &= (\mathbf{A}^\top\mathbf{A}+\lambda \mathbf{I})^{-1} [\mathbf{A}^\top(\boldsymbol{y} - (\mathbf{A}\mu_t + \mathbf{b})) + \mathbf{A}^\top\mathbf{A}\mu_t + \lambda \mu_t]\\
        &= (\mathbf{A}^\top\mathbf{A}+\lambda \mathbf{I})^{-1} [\mathbf{A}^\top(\boldsymbol{y} - \mathbf{A}\mu_t - \mathbf{b})] \\&\quad \quad \quad + (\mathbf{A}^\top\mathbf{A}+\lambda \mathbf{I})^{-1}(\mathbf{A}^\top\mathbf{A}+\lambda \mathbf{I})\mu_t\\
        &= (\mathbf{A}^\top\mathbf{A}+\lambda \mathbf{I})^{-1} \mathbf{A}^\top(\boldsymbol{y} - \mathbf{A}\mu_t - \mathbf{b}) + \mu_t
    \end{aligned}
\end{equation}
Next, we exploit the fact that the variance $\sigma_t$ of the denoising kernel in DDIM is very small as $t$ approaches zero, i.e., $\sigma_t \rightarrow 0$. For Eq. \ref{KKT} to hold, the multiplier $\lambda$ must be sufficiently large, i.e., $\lambda \rightarrow \infty$. In this asymptotic regime, the term $(\mathbf{A}^\top\mathbf{A}+\lambda \mathbf{I})^{-1}$ can be approximated as $\frac{1}{\lambda}\mathbf{I}$ (as $\mathbf{A}^\top\mathbf{A}$ becomes negligible compared to $\lambda \mathbf{I}$). Thus, the asymptotic form of $\boldsymbol{x}_t(\lambda)$ is:
\begin{equation}\label{eq6}
    \boldsymbol{x}_t(\lambda) \approx \frac{1}{\lambda} \mathbf{A}^\top(\boldsymbol{y}- \mathbf{A}\mu_t - \mathbf{b}) + \mu_t
\end{equation}
Substituting this result back into the constraint equation from Eq. \ref{KKT}:
\begin{equation}
    \left\| \left( \frac{1}{\lambda} \mathbf{A}^\top(\boldsymbol{y} - \mathbf{A}\mu_t - \mathbf{b}) + \mu_t \right) - \mu_t \right\|^2 = \sigma_t^2 d
\end{equation}
This simplifies to:
\begin{equation}
    \frac{1}{\lambda^2}\|\mathbf{A}^\top(\boldsymbol{y} - \mathbf{A}\mu_t - \mathbf{b})\|^2 = \sigma_t^2 d
\end{equation}
Solving for the multiplier $\lambda$:
\begin{equation}
    \lambda = \frac{\|\mathbf{A}^\top(\boldsymbol{y} - \mathbf{A}\mu_t - \mathbf{b})\|}{\sigma_t\sqrt{d}}
\end{equation}
Finally, we can find the asymptotic form of the optimal particle by substituting this expression for $\lambda$ back into Eq. \ref{eq6}:
\begin{equation}
    \begin{aligned}
        \boldsymbol{x}_t^*&\approx \mu_t + \left( \frac{\sigma_t\sqrt{d}}{\|\mathbf{A}^\top(\boldsymbol{y} - \mathbf{A}\mu_t - \mathbf{b})\|} \right)\mathbf{A}^\top(\boldsymbol{y} - \mathbf{A}\mu_t - \mathbf{b})\\
        &= \mu_t +\sigma_t\sqrt{d}\frac{\mathbf{A}^\top(\boldsymbol{y} - \mathbf{A}\mu_t - \mathbf{b})}{\|\mathbf{A}^\top(\boldsymbol{y} - \mathbf{A}\mu_t - \mathbf{b})\|}\\
        & (\text{Substituting } \mathbf{b}=\overline{\mathcal{H}}-\mathbf{A} \mu_t) \\
        & = \mu_t +\sigma_t\sqrt{d}\frac{\mathbf{A}^\top(\boldsymbol{y} - \overline{\mathcal{H}})}{\|\mathbf{A}^\top(\boldsymbol{y} - \overline{\mathcal{H}})\|}
    \end{aligned}
\end{equation}
The conclusion is confirmed.

\noindent \textbf{Intuitive Explanation:} This approximate solution has a very intuitive geometric interpretation:

1.  \textit{Starting Point:} The process begins at the center of the constrained sphere, $\mu_t$.

2.  \textit{Direction:} We move in the negative gradient direction of the objective function, calculated at the starting point $\mu_t$. The gradient of the objective function is $\nabla_{\boldsymbol{x}_t} = 2\mathbf{A}^T(\mathbf{A}\boldsymbol{x}_t+\mathbf{b}-\boldsymbol{y})$. At $\boldsymbol{x}_t = \mu_t$, the negative gradient direction is given by $\mathbf{A}^T(\boldsymbol{y} - (\mathbf{A}\mu_t + \mathbf{b}))$, which is the direction of steepest descent.

3.  \textit{Step Size:} The step size is a fixed distance, $\sigma_t\sqrt{d}$, which is determined by the radius of the constrained sphere.

In summary, when $\sigma_t$ is small, the optimal solution is approximately found by moving a fixed distance $\sigma_t\sqrt{d}$ along the steepest descent direction of the objective function, starting from the center of the sphere $\mu_t$.

\section{Appendix B: Experimental Setup}

\subsection{B1: Formulations of Inverse Problems}
\textbf{Image inpainting.} The formulation for our image inpainting problem follows recent work \cite{chung2022diffusion, song2023pseudoinverse}. The degradation model is represented as:
\begin{equation}
    \boldsymbol{y}=\mathbf{M} \odot \boldsymbol{x}+\boldsymbol{\eta}
\end{equation}
where $\mathbf{M}$is a binary mask with elements randomly set to 0 or 1, and $\odot$denotes element-wise multiplication. The goal of the inpainting task is to recover the missing pixels (where the mask is 0) from the observed pixels y. All measurements include additive Gaussian noise $\boldsymbol{\eta}$ with standard deviation $\sigma_y=0.05$.

\noindent \textbf{Super-resolution.} We consider bicubic super-resolution (SR) with the following degradation model:
\begin{equation}
    \boldsymbol{y}=\boldsymbol{x} \downarrow_{s f}^{\text {bicubic }}+\boldsymbol{\eta}
\end{equation}
where $\downarrow_{s f}^{\text {bicubic }}$denotes bicubic downsampling with a specified scale factor $s f$.

\noindent \textbf{Debluring.} The linear degradation model for image deblurring with Gaussian noise is expressed as:
\begin{equation}
    \boldsymbol{y}=\boldsymbol{x} \otimes \mathbf{k}+\boldsymbol{\eta}
\end{equation}
where $\otimes$ is a two-dimensional convolution operator and $\mathbf{k}$  is a blurring kernel applied across all image channels. 

\noindent \textbf{JPEG restoration.} This problem is formulated as:
\begin{equation}
    \boldsymbol{y}=h(\boldsymbol{x})+\boldsymbol{\eta}
\end{equation}
where $h(\boldsymbol{x})$ is the JPEG compression function. This function involves a discrete cosine transform (DCT) followed by a quantization step. We use the quantization matrix from  \citet{ehrlich2020quantization} to compress the original $256\times256$ image, along with $2\times2$ chroma subsampling. Note that in contrast to the setup in \citet{song2023pseudoinverse}, we introduce additional observation noise, making this task more challenging and realistic.

\begin{table*}[ht]
    \centering
\resizebox{1\textwidth}{!}{\begin{tabular}{lcccccc}
\toprule
     & Diffusion & Timestep & Guidance & Number & Number & Number of \\
    Method& Schedule & & Scale & of Update & of Particles & Forward Calls \\
\midrule
SCG \cite{huang2024symbolic} & VP \cite{song2020score} & 1000 & - & - & 64 & 64k \\
\midrule
DPG \cite{tang2024solving} & VP \cite{song2020score} & 1000 & 180.0 & 1 & 64 & 64k \\
\midrule
EnKG \cite{zhengensemble} & EDM \cite{karras2019style} & 100 & 2.0 & 2 & 320 & 64k \\
\midrule
\textbf{CPS (Ours)} & VP \cite{song2020score} & 500 & - & - & 64 & 64k \\
\bottomrule
\end{tabular}}
    \caption{ Hyperparameter configurations of the evaluated gradient-free methods for diffusion-based inverse problem solving. Hyperparameters that are not applicable to a specific method are denoted by a dash (–). For a fair comparison, we calibrated the number of particles for each method to ensure a consistent total count of forward observation model calls.}
    \label{tab:hyp}
\end{table*}

\noindent \textbf{Black Hole Imaging.} 
Black hole imaging measurements are obtained through very long baseline interferometry (VLBI) technology \cite{blackburn2020closure}. In this technique, each pair of telescopes $(a, b)$ provides a complex visibility measurement, $V_{a, b}^t$,by sampling specific spatial Fourier frequencies of the source image at time t. In practice, the visibility data is corrupted by three types of noise: telescope-based gain errors, telescope-based phase errors, and baseline-based thermal noise. The first two types of noise, caused by atmospheric turbulence and instrument miscalibration, are mitigated by combining multiple visibilities into noise-canceling data products: closure phases and log closure amplitudes. These are defined as:
\begin{equation}
    \begin{aligned}
& \boldsymbol{y}_{t,(a, b, c)}^{\mathrm{cph}}=\angle\left(V_{a, b}^t V_{b, c}^t V_{a, c}^t\right):=\mathcal{H}_{t,(a, b, c)}^{\mathrm{cph}}(\boldsymbol{x}) \\
& \boldsymbol{y}_{t,(a, b, c, d)}^{\mathrm{camp}}=\log \left(\tfrac{\left|V_{a, b}^t\right|\left|V_{c, d}^t\right|}{\left|V_{a, c}\right|\left|V_{b, d}^t\right|}\right):=\mathcal{H}_{t,(a, b, c, d)}^{\mathrm{camp}}(\boldsymbol{x})
\end{aligned}
\end{equation}
where $\angle$ computes the angle of a complex number. Additionally, the likelihood function typically includes a constraint on the total flux \cite{sun2024provable}. The overall likelihood for the forward model is then expressed by aggregating these data products over time and telescope combinations:
\begin{equation}
    \begin{aligned}
\ell(\boldsymbol{y} \mid \boldsymbol{x})= & \underbrace{\sum_{t, \mathrm{c}} \tfrac{\left\|\mathcal{H}_{t, \mathrm{c}}^{\mathrm{cph}}(\boldsymbol{x})-\boldsymbol{y}_{t, \mathrm{c}}^{\mathrm{cph}}\right\|_2^2}{2 \beta_{\mathrm{cph}}^2}}_{\chi_{\mathrm{cph}}^2}+\underbrace{\sum_{t, \mathrm{c}} \tfrac{\left\|\mathcal{H}_{t, \mathrm{c}}^{\mathrm{camp}}(\boldsymbol{x})-\boldsymbol{y}_{t, \mathrm{c}}^{\mathrm{camp}}\right\|_2^2}{2 \beta_{\mathrm{camp}}^2}}_{\chi_{\text {camp }}^2} \\
& +\rho \tfrac{\left\|\sum_i \boldsymbol{x}_i-\boldsymbol{y}^{\mathrm{flux}}\right\|_2^2}{2}
\end{aligned}
\end{equation}
where $\beta_{\mathrm{cph}}$ and $\beta_{\mathrm{camp}}$ assumed to be known standard deviations, $\boldsymbol{y}^{\text {flux }}$ is accurately measured, and $\rho>0$ is a parameter. The first two terms are often referred to as the $\chi^2$ errors. We conduct experiments using the dataset provided by \citet{zhenginversebench} and the checkpoint of the diffusion model. This includes 100 test images. We also consider the settings of three observation time ratios:
100\%, 10\% and 3\%.

\noindent \textbf{Fluid data assimilation.} The Navier-Stokes equation is a classic benchmark problem in fluid mechanics \cite{chandler2013invariant}. Its applications are wide-ranging, from ocean dynamics to climate modeling, and often involve calibrating the initial conditions of numerical forecasts by assimilating observational data.
\begin{equation}
    \begin{aligned}
\partial_t \boldsymbol{w}(\boldsymbol{x}, t)+\boldsymbol{u}(\boldsymbol{x}, t) \cdot \nabla \boldsymbol{w}(\boldsymbol{x}, t) & =\nu \Delta \boldsymbol{w}(\boldsymbol{x}, t)+f(\boldsymbol{x}),\\
\nabla \cdot \boldsymbol{u}(\boldsymbol{x}, t) & =\mathbf{0},\\
\boldsymbol{w}(\boldsymbol{x}, 0) & =\boldsymbol{w}_0(\boldsymbol{x}),\\
\boldsymbol{x} \in(0,2 \pi)^2&, t \in(0, T] 
\end{aligned}
\end{equation}
where $\boldsymbol{u}$ is the velocity field, $\boldsymbol{w}=\nabla \times \boldsymbol{u}$ is the vorticity, $\boldsymbol{w}_0$ is the initial vorticity, $\nu \in \mathbb{R}_{+}$is the kinematic viscosity coefficient, and $f(\boldsymbol{x})$ is the forcing function. We consider the inverse problem of recovering the initial vorticity field $\boldsymbol{w}_0$, from sparse and noisy measurements of the final state at $T=1$. The observation model is given by:
\begin{equation}
\boldsymbol{y}= S\mathcal{F}(\boldsymbol{w}_0)+\boldsymbol{\eta}    
\end{equation}
where $S$ is the sampling operator, $\boldsymbol{\eta} $ is the measurement noise, and $\mathcal{F}(\cdot)$ is the forward process of the numerical solver. We use a pseudo-spectral method \cite{li2020fourier} to solve the Navier-Stokes equations. Due to the inherent difficulty and numerical instability of backpropagation through such a solver, we treat the forward model as a black-box function. Our goal is to recover the 2D initial vorticity field from sparse and noisy measurements of size 128×128 at the final time $T=1$. We use a dataset provided by \citet{zhenginversebench} for testing, consisting of 10 different initial vorticity fields. We consider three settings to simulate sparse measurements, including 2×, 4×, and 8× downscaling. All measurements included noise with $\sigma_y=2.0$.

\subsection{B2: Implementation Details}
\textbf{Computing hardware and software:} All experiments, including computational time and memory usage measurements, were conducted on a single NVIDIA A100 GPU. The software stack consisted of Python 3.9 and PyTorch 2.4.

\noindent \textbf{Implementation of gradient-based method.} For the implementation of gradient-based diffusion inverse problem solvers, including DDRM \cite{kawar2022denoising}, DPS \cite{chung2022diffusion}, RED-diff \cite{chung2022diffusion}, and DAPS \cite{zhang2025improving}, we utilize the official repositories provided by their respective authors. Since an official implementation of $\Pi$GDM \cite{song2023pseudoinverse} was not available, we faithfully reproduce the method using the pseudo-code provided in the original paper. We adhere to the original hyperparameter configurations for all of these methods unless otherwise specified.

\noindent \textbf{Implementation of gradient-free method.} For our proposed CPS, we use a consistent set of hyperparameters across all tasks. We employ a 500-step stochastic DDIM solver \cite{song2020denoising} and maintain an ensemble of 64 particles for each step. To improve robustness, we incorporated the Restart strategy. Based on our observations that the effect of Restart diminishes in the later stages of the generation process, we applied the strategy only during the first 20\% of the sampling steps. We set the number of restarts, $N_r$, to 5, which improved the solution quality without a significant increase in computational burden. For a fair comparison with other gradient-free baseline methods (SCG, DPG, EnKG), we adjusted their particle counts to be consistent with the total number of forward model calls used by CPS. Our ablation studies in see main text show that the performance of these methods often improves with an increased number of particles. For consistency, we follow the default hyperparameter settings from the official implementations of these methods, with the exception of particle count. Specific parameter settings are summarized in Table~\ref{tab:hyp}.

The official code repositories for the competing methods used in our evaluation are listed below:

\begin{itemize}
\item DDRM \cite{kawar2022denoising}: \url{https://github.com/bahjat-kawar/ddrm}
\item DPS \cite{chung2022diffusion}: \url{https://github.com/DPS2022/diffusion-posterior-sampling}
\item RED-diff \cite{mardani2023variational}: \url{https://github.com/NVlabs/RED-diff}
\item DAPS \cite{zhang2025improving}: \url{https://github.com/zhangbingliang2019/DAPS}
\item SCG \cite{huang2024symbolic}: \url{https://github.com/yjhuangcd/rule-guided-music}
\item DPG \cite{tang2024solving}: \url{https://github.com/loveisbasa/DPG}
\item EnKG \cite{zhengensemble}: \url{https://github.com/devzhk/enkg-pytorch}
\end{itemize}

\noindent \textbf{Model checkpoints and datasets.} For the image inverse problems, we conducted our experiments on the benchmark dataset FFHQ \cite{karras2019style}. We use a pretrained diffusion model provided by \citet{chung2022diffusion} and randomly sample 100 images from this dataset to compute our reported metrics. For the two scientific inverse problems, black hole imaging and fluid data assimilation, we utilize InverseBench \cite{zhenginversebench}, a dedicated physical science inverse problem benchmark. InverseBench provides both the relevant pre-trained models and test data used in our experiments.
\section{Appendix C: Additional Results}

\begin{figure}[htbp]
     \centering
\includegraphics[width=\columnwidth]{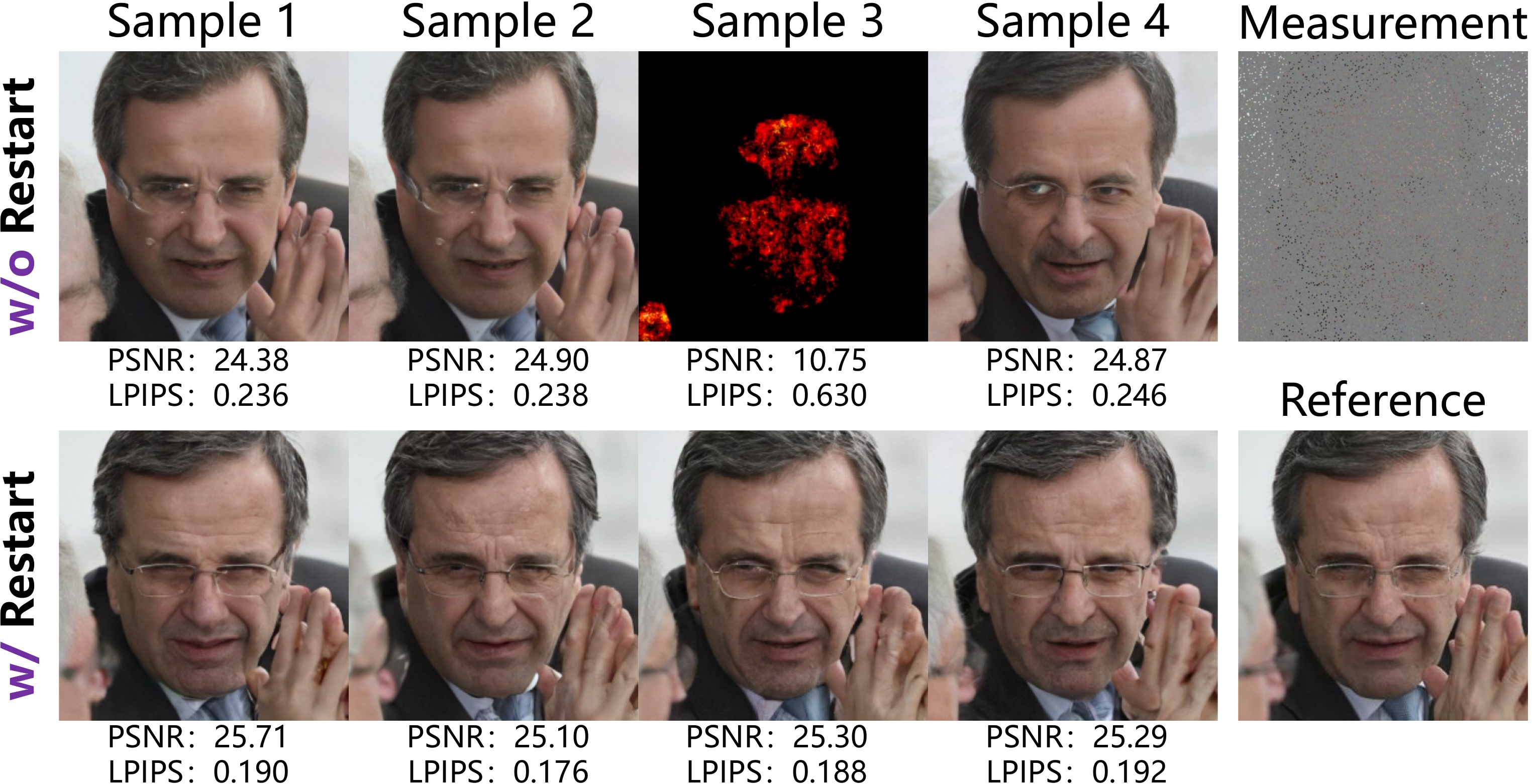}
    \caption{Ablation study of the Restart strategy with different random seed initializations.}
    \label{restart_fig}
\end{figure}

Figure~\ref{restart_fig} shows the impact of the Restart strategy in CPS, using different random seeds to initialize the noise.

\section{Appendix D: Discussion and Limitation}
Our work introduces \textbf{\emph{Constrained Particle Search (CPS)}}, a new gradient-free method that addresses the performance of existing inverse problem solvers with diffusion models. It works well even with only a small number of particles sampled. A core philosophical contribution of our work is the shift from passive candidate selection to active particle search. Previous methods, such as SCG, would discard most of the candidate particles at each step, resulting in computational waste. Our empirical research findings indicate that even seemingly 'bad' particles contain useful information, providing a strong impetus to move beyond this rejection sampling paradigm. By redefining the problem as a constrained optimization task, CPS effectively integrates information from all candidates, thereby achieving a more efficient and robust sampling process.

Although CPS has made significant progress in solving gradient-free diffusion inverse problems, it is not without limitations, which provide avenues for future research. CPS still requires sampling multiple particles at each step to estimate the surrogate model. This increases the computational cost compared to unconditional sampling. One potential direction is to combine faster solvers \cite{xue2023sa} or shortcut sampling \cite{liu2023accelerating}, starting from initial guesses rather than noise, to further improve the efficiency of our framework.

\end{document}